# MI CAM: Mutual Information Weighted Activation Mapping for Causal Visual Explanations of Convolutional Neural Networks

Ram S Iyer
Rajiv Gandhi Institute of Petroleum Technology
Jais, 229304, Uttar Pradesh, India
ramsiyer77@gmail.com

Narayan S Iyer
National Institute of Technology Rourkela
Rourkela, 769008, Odisha, India
narayansiyer7@gmail.com

Rugmini Ammal P
ZGC Calicut
Calicut, 673014, Kerala, India
rugmini75@gmail.com

## Abstract

*With the intervention of machine vision in our crucial day to day necessities including healthcare and automated power plants, attention has been drawn to the internal mechanisms of convolutional neural networks, and the reason why the network provides specific inferences. This paper proposes a novel post-hoc visual explanation method called MI CAM based on activation mapping. Differing from previous class activation mapping based approaches, MI CAM produces saliency visualizations by weighing each feature map through its mutual information with the input image and the final result is generated by a linear combination of weights and activation maps. It also adheres to producing causal interpretations as validated with the help of counterfactual analysis. We aim to exhibit the visual performance and unbiased justifications for the model inferencing procedure achieved by MI CAM. Our approach works at par with all state-of-the-art methods but particularly outperforms some in terms of qualitative and quantitative measures. The implementation of proposed method can be found on* https://anonymous.4open.science/r/MI-CAM-4D27

## 1. Introduction

The advent of deep neural networks (DNNs) and convolutional neural networks (CNNs) in particular, combined with new innovative neural net architectures are producing constant outbreaks in various computer vision tasks ranging from image classification [1-4] , object detection [5-7] and image segmentation to advanced synthetic image generation, image captioning, visual question answering and 3D visual data analytics [8-11] . Though these networks, without a doubt, provide remarkable performances, their lack of decomposability [12] , complex architecture and black box nature render them as hardly interpretable. Hence, the field of explainable artificial intelligence (XAI), with special focus on image based models is earnestly studied in this particular context in order to obtain enhanced explanations for model inferences. This aims to increase the model transparency leading the path towards grey-box models [13] .

Regarding methods for model explanation, widely adopted methods involve Gradient based [14] , Perturbation based [15] , Decomposition based [16-18] and Class Activation Map (CAM) based [19-21] visualizations. Gradient-based methods propagate the gradient of a target class back to the input layer to emphasize the image regions that strongly affect the prediction, where as, perturbation-based approaches analyse the change to prediction of the model caused by perturbing original input, for model explanation. Decomposition-based techniques decompose the predictions into contributions from each feature in order for model explainability. CAM-based method explains model decision by generating a weighted linear combination of the convolutional feature maps to assert important regions.

### 1.1. Related Works in CAM

The initiation for a CAM based visualisation method was put forward by B Zhou et al. [22] in his work. The vanilla CAM visualizes class specific regions using a linearly weighted combination of the last convolutional layer feature maps. RR Selvaraju et al. [23] developed Grad-CAM, a further approach to CAM, by leveraging the gradient information of the predicted class with respect to each feature map as the weighting parameter. Grad-CAM ++

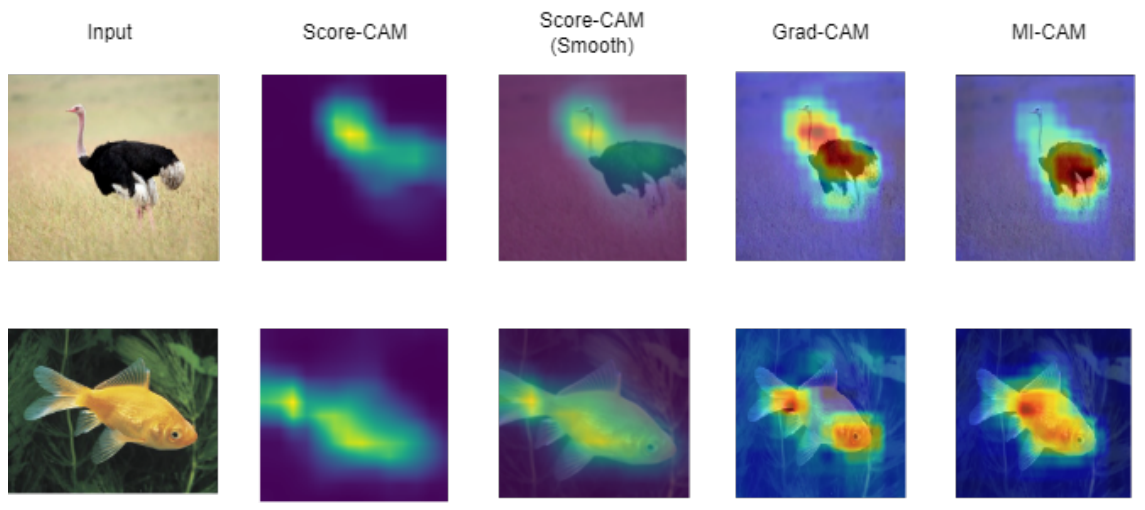

Figure 1. Visual comparison of MI-CAM with related works Score-CAM, Score-CAM (Smooth) and Grad-CAM. MI-CAM provides high resolution visualizations.

[24] and Integrated Grad-CAM [25] are further works and advances on gradient based CAM.

Earlier works on gradient-free CAM visual explanation approaches was done in Score-CAM by H Wang et al. [26] . It introduces a gradient-free visual explanation method by bridging the gap between perturbation based and CAM based methods. Modifications to Score-CAM was done by Yifan Chen and Guoqiang Zhong in their work on Score-CAM ++ [27] by introducing a feature map selection procedure prior to CAM generation. Similarly, by introducing the integration operation within the Score-CAM pipeline, Integrated Score-CAM [28] was formulated by R Naidu et al. Principal Components (PC) was used to calculate the weight values of the corresponding feature maps through eigen value decomposition in Eigen-CAM by Mohammed Bany et al. [29] .

However, though these explanation methods have evolved a lot with each progression, each methods have the limitations of their own. The vanilla CAM method requires a max pooling layer to follow the convolutional layer of interest in order to be applicable. This shows the architecture-sensitivity of the method and the heavy constraints put forward by model architecture on the said method. Similarly the gradient based methods of Grad-CAM and its derivatives incur gradient based issues such as the shattered gradient problem [30] , vanishing / exploding gradient problems and saturation of gradients as the network deepens. These lead to noisiness, discontinuity and false confidence in the visualization. Score based and decomposition based CAM methods also have certain short comings including computational complexity, as shown in supplementary material, due to the multiple forward passes required to compute the importance of each activation map. It depends extensively on model outputs, becoming sensitive to adversarial examples. On the note, apart from this, all the above referred activation mapping methodologies have neither analyzed nor show guarantee of causality in the explanations generated [31-33] .

### 1.2. Our Contributions

Gaining momentum from the existing CAM based visualization methods, we revisit these approaches and discuss our concerns of how they might be providing biased visualizations and erase latent information. In an attempt to address these hurdles, we propose a novel post-hoc visual explanation method, named MI CAM. MI CAM uses the mutual information between feature maps and input images to weight the maps, producing improved causal visualizations. We summarize our contributions as:

1. We propose a novel CNN visual explanation method named MI CAM, bridging the gap between information theory and CAM-based explanation methods and representing the weights of activation maps in an information-gain centric manner.

2. We evaluate the fairness and effectiveness of generated weighted activation maps across layers and on recognition tasks, quantitatively through Average Drop / Average Increase and Deletion/Insertion Curves and qualitatively through visual inspection.

3. We demonstrate the causality of the generated heatmaps with the help of counterfactual analysis. MI CAM shows remarkable performance in providing causal representations and showcasing latent properties essential for effective CAM-based visual explanations.

## 2. Background

We first introduce the notations being considered in this paper, prior to a background study detailing the prerequisites and causes of the current research.

**Notation:** Let $Y = f(X)$ be a convolutional neural network, taking an input $X \in R^d$ to produce a probability distribution $Y$, such that $l$ denotes a given layer. For a given layer, $A_l$ and ${A_l}^k$ represent the activation maps for the $l$-th layer and the $k$-th channel of that particular layer, in specific.

### 2.1. CAM

As discussed earlier, class activation mapping (CAM) is an explanation method for identifying discriminative regions by linearly weighted combination of the outputs of the last convolutional layer before the global average pooling layer. The weights being devoted to each of the global average pooling outputs by the channel corresponding to the class of interest in the following fully connected layer serves as the weights of the feature maps to provide the activation map.

**Definition 1:** Let a CNN model $f$ having a global average pooling layer $l$ and a convolutional layer *l-1* and a fully connected layer *l+1* preceding and following it respectively. Given a class of interest $c$, vanilla CAM can be expressed as

$$L^c_{CAM} = ReLU(\sum_k \alpha^c_k A^k_{l-1}) \quad (1)$$

where

$$\alpha^c_k = w^c_{l,l+1}[k] \quad (2)$$

with $w^c_{l,l+1}[k]$ being the weight between the *l* and *l+1* layer for $k$-th neuron. Since this is only feasible in architectures where the final convolutional layer is followed by a global average pooling layer and further, by the final fully connected layer, this method is not generalizable across varying architectures or at least not without re-training the data on the altered architecture. In order to cope up with this barrier, the weight values of the activation maps were modified to

$$\alpha^c_k = GAP(\frac{\partial Y^c}{\partial A^k_l}) \quad (3)$$

with GAP(.) being the global average pooling operation.

### 2.2. Causal Explanations

**Correlation vs Causation:** Causality in the explanations, as discussed earlier, is a major factor that emphasizes the quality of a CAM approach. Weighting the feature maps through correlational metrics may not suffice the needs of an explainability method. It shows which part of the input contribute most to the activation of specific class, but doesn't explicitly model or measure causal relationships, i.e. how much information a feature provides about the input. Major share of CAM methods have not considered the evaluation of causality of the produced explanations. In a neural network context, the output is a function of the input through a series of transformations across layers. Each layer generates feature maps $A^k_l$, which can be seen as intermediate causes that contribute to the final output.

**Definition 2:** A model can be thought of as a composition of functions

$$Y = f(X) = f_L(f_{L-1}(...(f_2(f_1(X))...)) \quad (4)$$

and the feature map at layer $l$ can be represented as

$$A^k_l = f^{(k)}_l(f_{l-1}(...f_2(f_1(X))...)) \quad (5)$$

The causal relationship can be expressed in terms of the joint probability distribution

$$P(A^k_l, Y) = P(Y|A^k_l) \cdot P(A^k_l) \quad (6)$$

This indicates that the output $Y$ depends on the feature map $A^k_l$ and we can factorize the joint distribution accordingly. Therefore the causal effect estimate of a feature map on the output can be denoted using do-calculus as

$$P(Y \mid do(A^k_l = a)) \quad (7)$$

### 2.3. Mutual Information

Mutual information, as mentioned earlier, is a dependence metric that quantifies the information contained by a random variable regarding another random variable. Let two random variables be $R_1$ and $R_2$ and their corresponding entropy be $H(R_1)$ and $H(R_2)$, such that

$$H(R_1) = -\sum_{r_1} P(r_1) \log P(r_1) \quad (8)$$

$$H(R_2) = -\sum_{r_2} P(r_2) \log P(r_2) \quad (9)$$

$$H(R_1, R_2) = -\sum_{r_1}\sum_{r_2} P(r_1, r_2) \log P(r_1, r_2) \quad (10)$$

where $P(r_1)$ and $P(r_2)$ are the marginal probabilities of $R_1 = r_1$ and $R_2 = r_2$ respectively and $P(r_1, r_2)$ is the joint probability. The mutual information between the two random variables is:

$$I(R_1; R_2) = H(R_1) + H(R_2) - H(R_1, R_2) \quad (11)$$
$$= \sum_{r_1}\sum_{r_2} P(r_1, r_2) \log \left(\frac{P(r_1)P(r_2)}{P(r_1, r_2)}\right)$$

where $H(R_1, R_2)$ is the joint entropy of $R_1$ and $R_2$.

## 3. MI CAM: Proposed Approach

This section describes the functioning of the proposed MI CAM for explaining the regions of focus being emphasized by the CNN-based model. Before MI CAM, previously explored methods analyzed the activation maps of the intermediate layers as a means of quantifying the efficiency of the explainability method. However, according to the conditional independence property of causality,

**Definition 3:** Given a CNN model $Y = f(X)$ having layers $l$, $l+1$ and $l-1$, producing activations $A^k_l$, $A^k_{l+1}$ and $A^k_{l-1}$ respectively as in Fig 2.

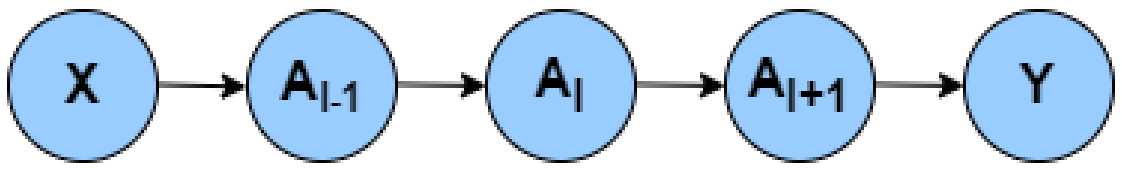

Figure 2. Directed Acyclic Graph (DAG) for the causal representation of a neural network. The arrow indicates the cause and the outcome.

Here, input $X$ causes the activation $A^k_{l-1}$ by putting some information in to it and according to conditional independence, this makes $A^k_{l-1}$ to contain all the information about $X$ that is relevant to determine the value of $A^k_l$, i.e. knowing the value of $A^k_{l-1}$ makes $X$ irrelevant for $A^k_l$. This can

be expressed as

$$P(A_l^k|A_{l-1}^k, X) = P(A_l^k|A_{l-1}^k) \tag{12}$$

and it is known as $A_{l-1}^k$ blocks $X$ from $A_l^k$ or $A_{l-1}^k$ d-seperates [34] $X$ and $A_l^k$. This same ideology can be propelled across the entire layers of the CNN. In summary, knowing the information passed on by the activations of the last layer encompasses the information passed on by all the preceding layers in itself.

## 3.1. Methodology

We now introduce the methodology of our proposed technique as per the pipeline illustrated in Fig 4. In contrast to earlier methods that heavily relies on the predicted class for weighting the importance of each activation map and explaining the model's decision making process, we incorporate the mutual information between each activation map and the input as the importance metric.

**Definition 4:** Taking the same CNN model $Y = f(X)$ as earlier, we pick the last convolution layer $l$ with $N$ channels. For the probability distribution estimations of the activation $P(A_l^k)$ and the input $P(X)$,

$$P(A_l^k) = P(\ flat(\ Up(\ A_l^k))) \tag{13}$$

$$P(X) = P(\ flat(\ Gr(\ X))) \tag{14}$$

where $Up(.)$ denotes the up-sampling operation of $A_l^k$ in to the input size and $flat(.)$ is the flattening function that maps the 2D matrix into a 1D vector and $Gr(.)$ is the gray-scale operation done on input. The joint probability is $P(A_l^k, X)$. Now, the mutual information measure between $A_l^k$ and $X$ is obtained as

$$I(A_l^k; X) = \sum_{a \in A_l^k} \sum_{x \in X} P(A_l^k = a, X = x) \times \log\left(\frac{P(A_l^k = a, X = x)}{P(A_l^k = a)P(X = x)}\right) \tag{15}$$

**Importance of Upsampling:** The activation maps are initially upsampled to the size of the input before undergoing flattening and mutual information calculation. This stands important as mutual information requires pairwise correspondence between two sets of observations, i.e. each element in the first vector should correspond to exactly one element in the second vector. Further, for the joint probability calculation, there is a need for a consistent number of samples in both the variables $A_l^k$ and $X$ necessitating vectors of same length.

Finally we define our proposed visual explanation method MI CAM in Def.5. The complete implementation details of the same is described in Algorithm 1.

**Definition 5 (MI CAM):** With the notations stated earlier, considering a convolutional layer $l$ of a model $f$, given an input $X$, MI CAM $L_{MI\ CAM}$ can be defined as

$$L_{MI\ CAM} = ReLU(\sum_k \alpha_k A_l^k) \tag{16}$$

where

$$\alpha_k = I(A_l^k; X) \tag{17}$$

where $I(.)$ is the mutual information score.

MI CAM gets rid of the dependence on gradients and eliminates class-specific biases. It aligns to providing causal explanations encapsulating both linear and non-linear dependencies. Although the last convolution layer is the one of primary focus, as it encapsulates all the information till there, as in Def.3, any intermediate convolution layer can be used as a possible target of our proposed framework.

**Algorithm 1 :** MI CAM Algorithm

**Require:** Input Image $X$, Model $f(X)$, layer $l$, number of channels $n$
**Ensure:** $L_{MI\ CAM}$
initialization
1. Get activation of layer l.
$A_l \leftarrow f_l(X)$
**for** $k\ in\ n$ **do**
2. $A_l^k \leftarrow Up(A_l^k)$ Up(.) - Upsampling
$X \leftarrow Gr(X)$ Gr(.) - Gray-scaling
3. Flatten activation map and input
$A_l^k \leftarrow flat(A_l^k)$, $X \leftarrow flat(X)$
4. Calculate mutual information
$I(A_l^k; X)$
**end for**
5. $L_{MI\ CAM} \leftarrow ReLU(\sum_k \alpha_k A_l^k)$

# 4. Experiment

Here we evaluate the effectiveness of our proposed method through certain experiments. We start with the qualitative evaluation of our method through visualization on test instances. Then we evaluate the faithfulness of the proposed method through Average Drop (AD), Average Increase (AI) and Area Under Deletion/Insertion Curve (AUC) metrics. Finally we evaluate the causality of the proposed method through counterfactual analysis. Above experiments are done on the VGG-16 network [35] pretrained on ImageNet, unless stated otherwise.

## 4.1. Visual Analysis for Qualitative Assessment

We provide a qualitative comparison of the heatmaps generated by various state of the art methods, including gradient-based, score-based and eigenvalue-based CAM approaches. Our method excels in producing visually assuring heatmaps, signifying the actual spatial regions being focused by the model with less random noise representations. We summarize our results in Fig 4. As visible, the random noises in MI CAM are less than other CAM methods of Grad-CAM, Grad-CAM++, Score-CAM and Eigen-CAM.

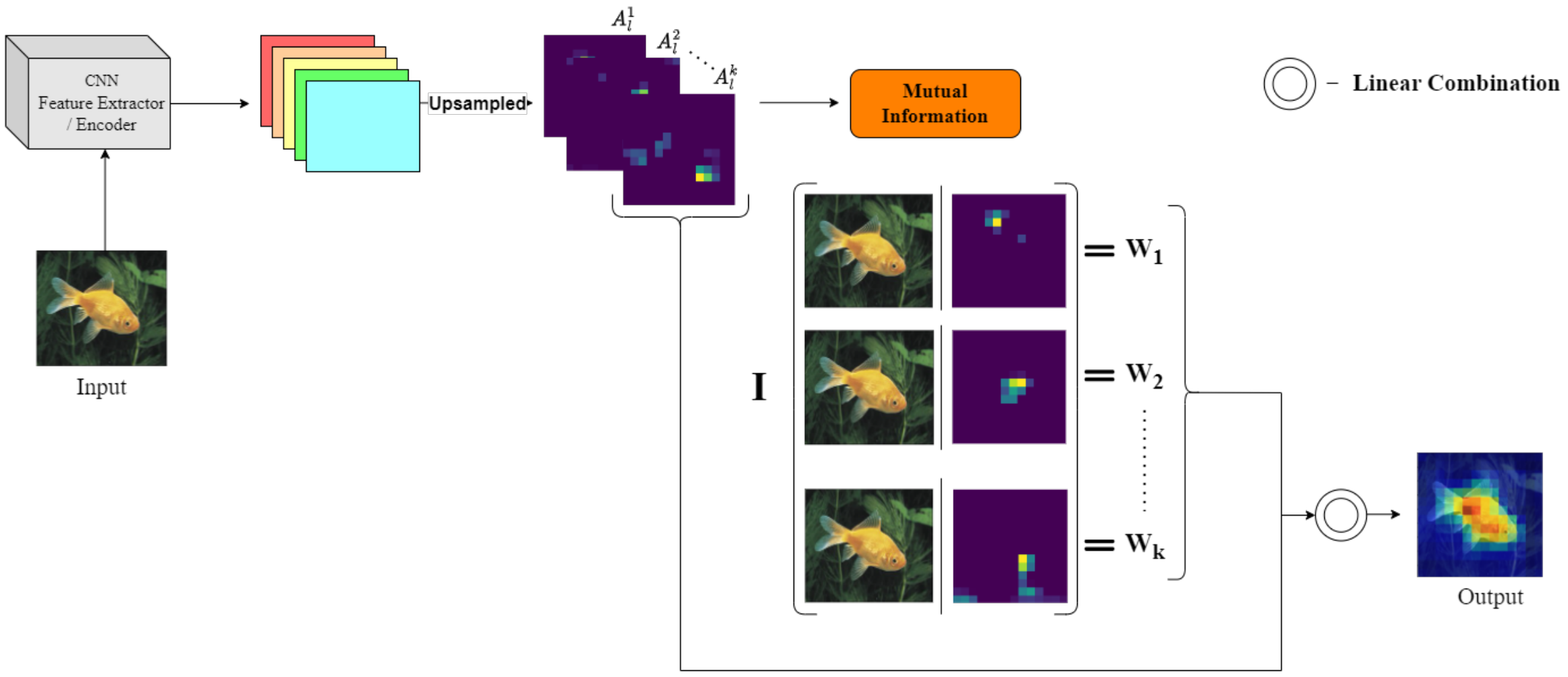

Figure 3. Pipeline of the proposed MI CAM. The input image is fed into any CNN feature extractor or encoder irrespective of the head network to extract the activation maps. The mutual information of the input image with each activation is calculated to obtain $k$ mutual information values corresponding to $k$ activation maps. These values are then used as the weighting parameters to produce the final results by linear combination of these weights and the activation maps.

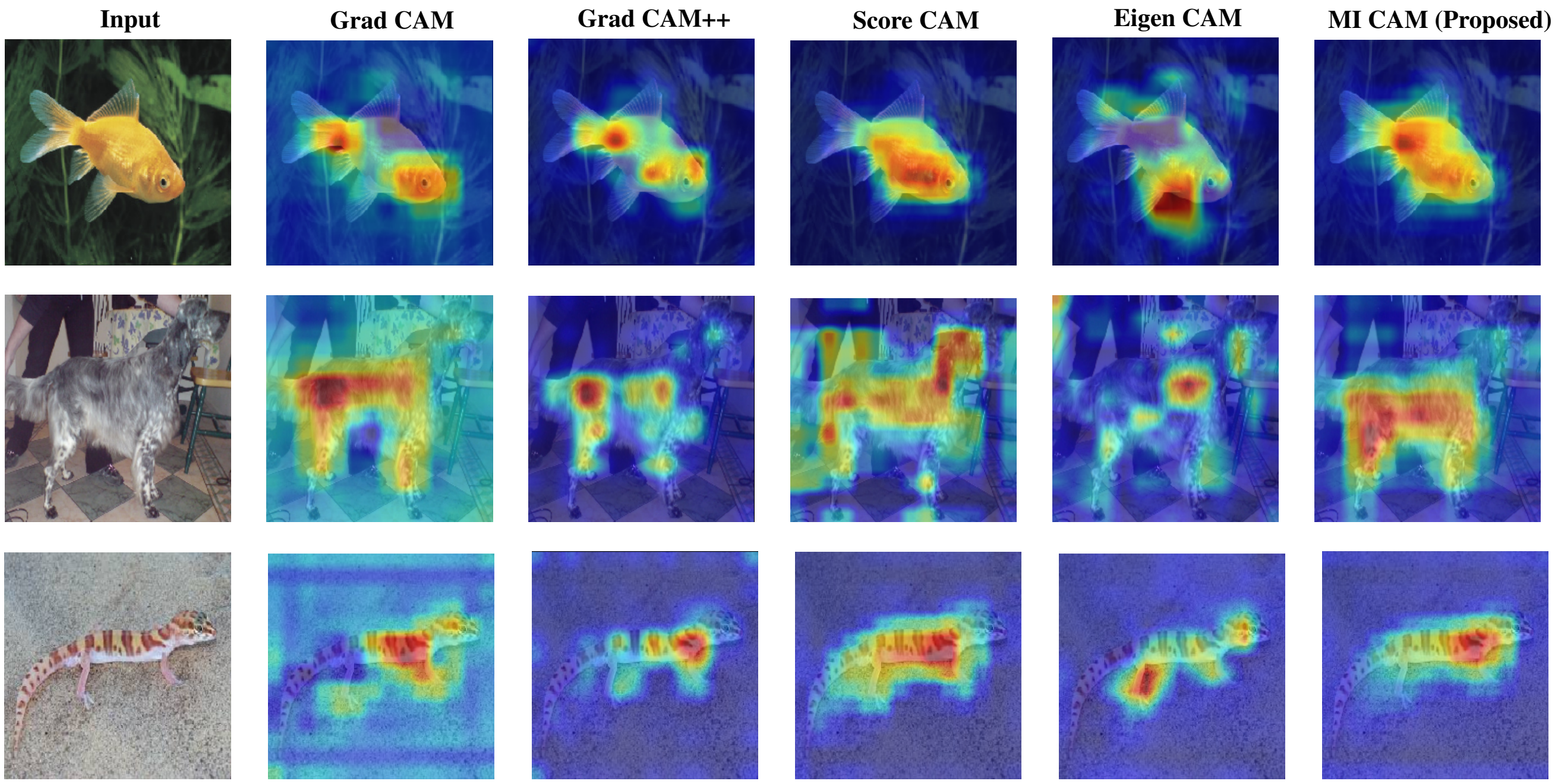

Figure 4. Visualization results of certain state of the art CAM based explanation methods Grad-CAM, Grad-CAM++, Score-CAM, Eigen-CAM and our proposed MI CAM.

### 4.2. Multi-Target Visualization

Holistic localization of multiple objects of the same class in an image can be attributed as another strong feature of MI CAM in line with the earlier CAM methodologies.

Fig 5 depicts the comparative results of others' and our approaches. Earlier approaches do locate both objects to an extend but lack the required focus on any one object. In MI CAM high focus (red portion) is distributed evenly to all the objects in the image.

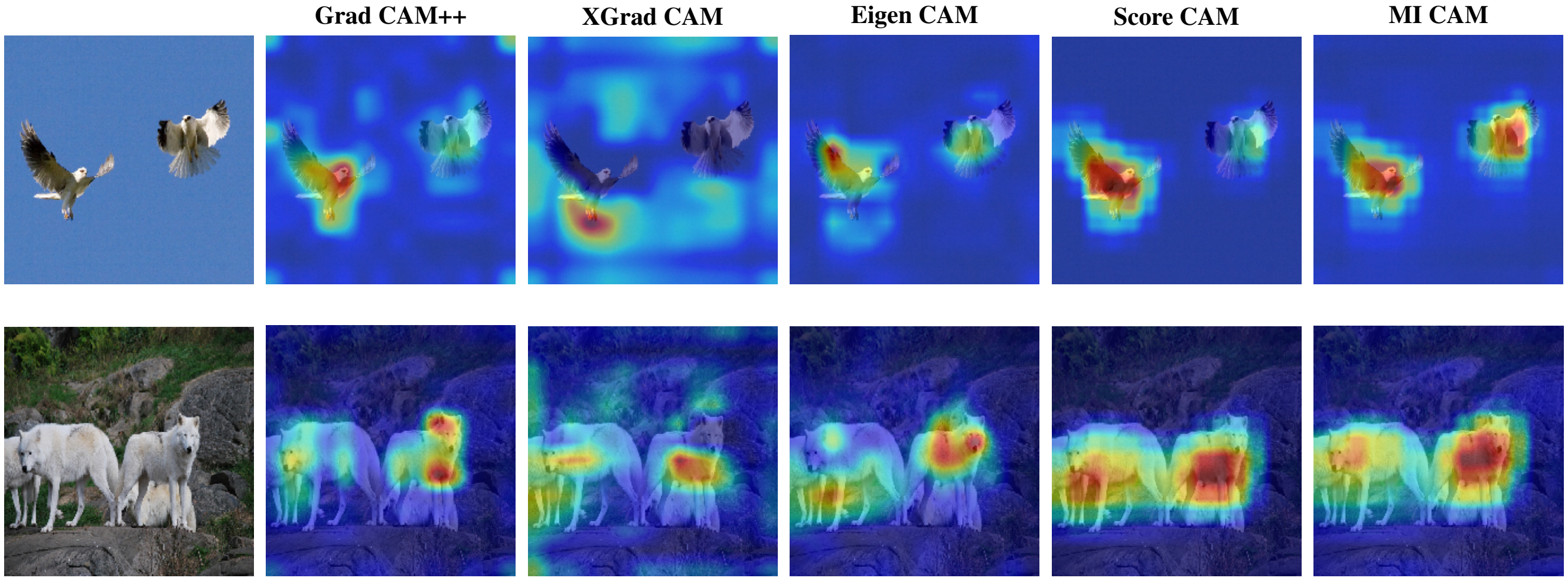

Figure 5. Results on images with multiple objects. MI CAM further capitalizes on other compared methods to cover every important spatial locations to provide a holistic explanation of all objects in the image

As each activation map is weighted by its mutual information with the input, each object irrespective of the contribution to target score, if provides sufficient information regarding the input, will be highlighted. Therefore, all objects that are deemed informative by the model for its functioning can be brought out.

### 4.3. Faithfulness via Quantitative Evaluation

The faithfulness of the explanation generated by MI CAM is analyzed using Average Drop (AD), Average Increase (AI) and Area Under Deletion/Insertion Curve (AUC) metrics.

The genuine inputs are perturbed or masked pixel-wise with the generated class activation mapping pixels in a one-to-one manner following a certain conditioning. This is done by point-wise multiplication in earlier studies. Here, the condition is altered a bit.

For AD calculation, the pixels of the generated activation mapping are normalized and ranked based on their intensity values. Further, a threshold is set and all the pixel values of the input corresponding to pixels below the threshold in the activation map are muted out. The class score on the masked input is calculated and the drop in confidence is quantified, in an attempt to obtain the relevance of the locations being emphasized by the proposed method.

Similarly, for AI, a baseline input (all pixel values set to '0' here) is fixed and pixels with highest intensity values, above a given threshold, are added to the baseline input. The difference in probability scores on the baseline sample and the masked sample is calculated to determine the increase in confidence obtained by introducing pixels deemed to be of highest importance. In both the experiments the threshold was set at 0.5.

The AD as expressed as $\sum_{i=1}^{N} \frac{max(0, Y_i^c - O_i^c)}{Y_i^c} \times \frac{100}{N}$, where $Y_i^c$ is the predicted score for class $c$ on actual image and $O_i^c$ is the predicted score on CAM-masked images and AI as expressed as $\sum_{i=1}^{N} \frac{max(0, O_i^c - B_i^c)}{O_i^c} \times \frac{100}{N}$, where $O_i^c$ and $B_i^c$ denote the predicted scores for class $c$ on input with CAM-focused pixels and baseline input respectively, are incorporated to evaluate the faithfulness of the method. Since muting pixel values with less relevance should not affect the probability scores much, the lower the AD, the better will be the CAM approach. Similarly, the higher the AI, the better will be the CAM approach. $N$ is the total number of input samples from the ImageNet (ILSVRC2012) validation data subset.

The results of the above evaluation in this setup for our proposed MI CAM and the other CAM approaches are summarized in Table 1. According to Table 1, MI CAM attains an AD of 79.02% and an AI of 43.89%, surpassing the performance of other CAM-based methods.

Area under deletion/insertion curves are other quantitative metrics intended to showcase the effectiveness of a CAM approach. While deletion curve plots the fraction of pixels deleted against the predicted probability scores, insertion curve graphs the fraction of pixels inserted against the probability scores. The pixels recognized as of highest relevance by the CAM method are arranged and they are removed from the image in certain ratio per step, starting from the most important first. The prediction score of the class of interest on the image is obtained simultaneously per step to plot the deletion curve. The area under the deletion curve

| Method | Avg. Drop (AD) | Avg. Increase (AI) |
|---|---|---|
| Grad CAM | 90.3937 | 39.0610 |
| Grad CAM++ | 94.9783 | 35.1189 |
| Score CAM | 86.8746 | 40.7467 |
| XGrad CAM | 92.7421 | 40.6317 |
| Eigen CAM | 94.6549 | 15.6492 |
| MI CAM | **79.0278** | **43.8921** |

Table 1. Evaluation results for AD and AI for VGG-16. Lower AD and Higher AI signifies a better performance of the explainability method.

(AUC) quantifies the decrease in class probability score incurred by the input image after the removal of the pixels indicated by the CAM method and hence a lower AUC of deletion curve suggests better CAM approach. Similarly, the important pixels are arranged in an ascending order and a certain ratio of pixels are inserted, starting from least important, and the corresponding class scores are obtained. This, plotted in the manner discussed above, forms the insertion curve. The more higher the AUC of insertion curve, the more robust the CAM method is.

| | Grad CAM | Grad CAM++ | Score CAM | XGrad CAM | Eigen CAM | MI CAM |
|---|---|---|---|---|---|---|
| Deletion | 0.36449 | 0.40690 | 0.35470 | 0.35909 | 0.35603 | **0.33424** |
| Insertion | 0.01047 | 0.00151 | 0.01124 | 0.00767 | 0.00755 | **0.01347** |

Table 2. Comparative analysis report of MI CAM and other approaches in terms of average AUC of deletion curve (lower is better) and insertion curve (higher is better) for VGG-16.

Table 2 reports the average AUC of deletion and insertion curves of MI CAM on VGG-16 to comprehend the faithfulness of the proposal. We remove or insert the pixels by setting their values to 0 or pixel intensity respectively [26] and 1% of pixels of the whole image are deleted or inserted per step for a total of 100 steps. Better performance of our method on the deletion curve and insertion curve metrics can be inferred from Table 2.

### 4.4. Localization Evaluation

The quality of the generated saliency maps and its ability to be applicable to weakly supervised localization and segmentation tasks is measured in this section through the localization evaluation. We conduct quantitative assessment of localization through pointing game and energy-based pointing game (EBPG) [26] metrics. Pointing game signifies the frequency with which the pixel with highest importance as shown by the CAM approach lies inside the bounding box annotation in the set of test samples and EBPG signifies the ratio of pixels with highest importance within the bounding boxes for a sample.

Table 3 shows the pointing game and EBPG results for MI CAM and other CAM methods.

| Method | Pointing Game (hit-rate) | EBPG (%) |
|---|---|---|
| Grad CAM | 0.60 | 53.714 |
| Grad CAM++ | 0.60 | 54.187 |
| Score CAM | 0.56 | 54.905 |
| XGrad CAM | 0.53 | 52.438 |
| Eigen CAM | **0.76** | 54.164 |
| MI CAM | 0.70 | **55.654** |

Table 3. Quantitative localization results for pointing game and EBPG for VGG-16. Higher hit-rate and EBPG % signifies a better localization performance of the explainability method.

It can be inferred that Eigen CAM obtained the highest hit-rate in pointing game followed by MI CAM, having hit-rates of 0.76 and 0.7 respectively. Impressively, MI CAM displays the best performance in terms of EBPG% for localization task with a EBPG% of 55.654%. Visual results of localization are included along with in Fig 6 to give further insight into how MI CAM can become useful in weakly supervised localization.

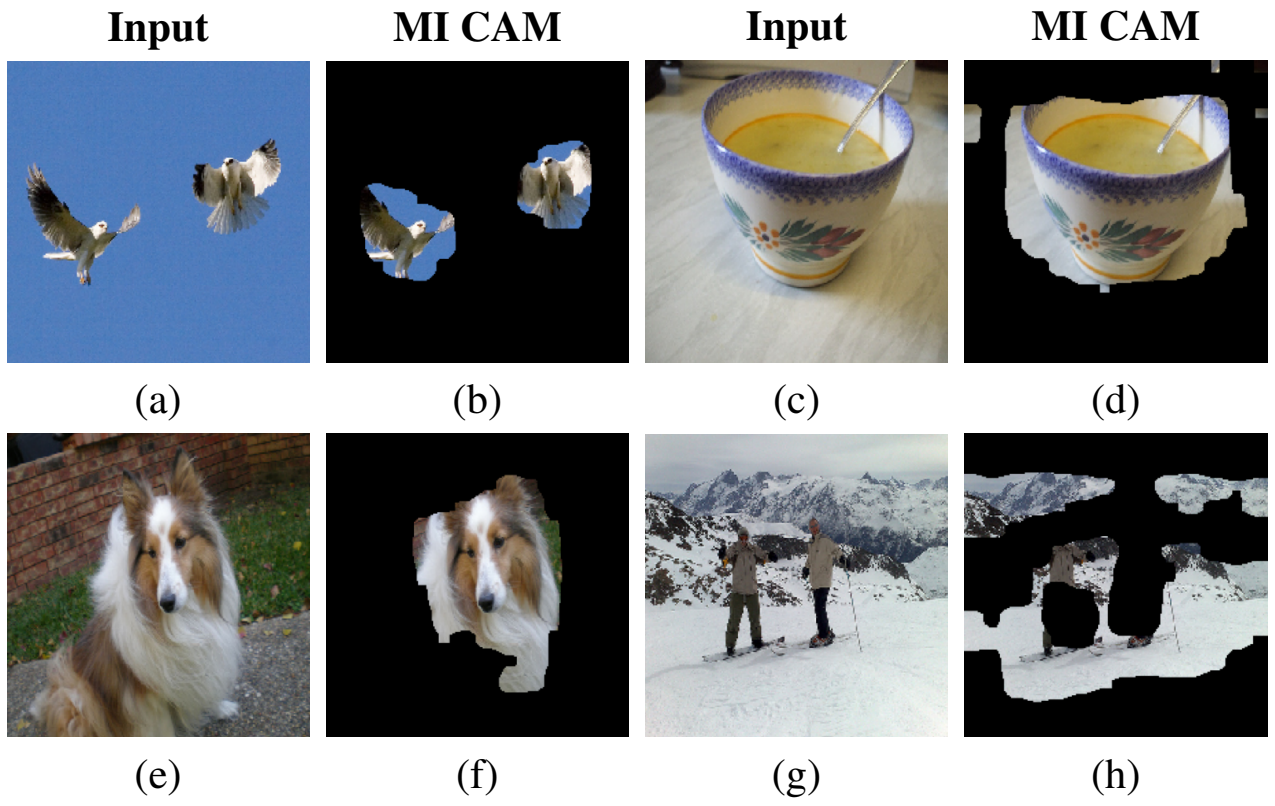

Figure 6. Visualization of localization results in weakly supervised scenario. (a), (c), (e) and (g) are kite bird, soup bowl, Shetland dog and Alp mountain respectively and (b), (d), (f) and (h) are corresponding localizations

It can be inferred from Fig 6 that MI CAM even localizes multiple object in (b) and heavily occluded Alps mountains situated in the background in (h) with excellent precision. This showcases its ability to be of use in erasing unwanted objects from an important context in the input.

### 4.5. Counterfactual Analysis

As cited in [32] and the earlier sections of the manuscript, it is important to ensure that the saliency maps generated by the explainability method are causal explanations in order for them to be robust and to take both object and contextual information into consideration. As per equation (7) in Def. 2, counterfactual examples can be generated from the genuine examples by setting or perturbing certain pixels to particular values. The weights of the feature maps generated from both the actual images and counterfactual images should display variations for the explanations to be causal. Fig 7 displays the counterfactual analysis results for some test samples. It can be checked that there are major variations in the mutual information weight of both actual and counterfactual images.

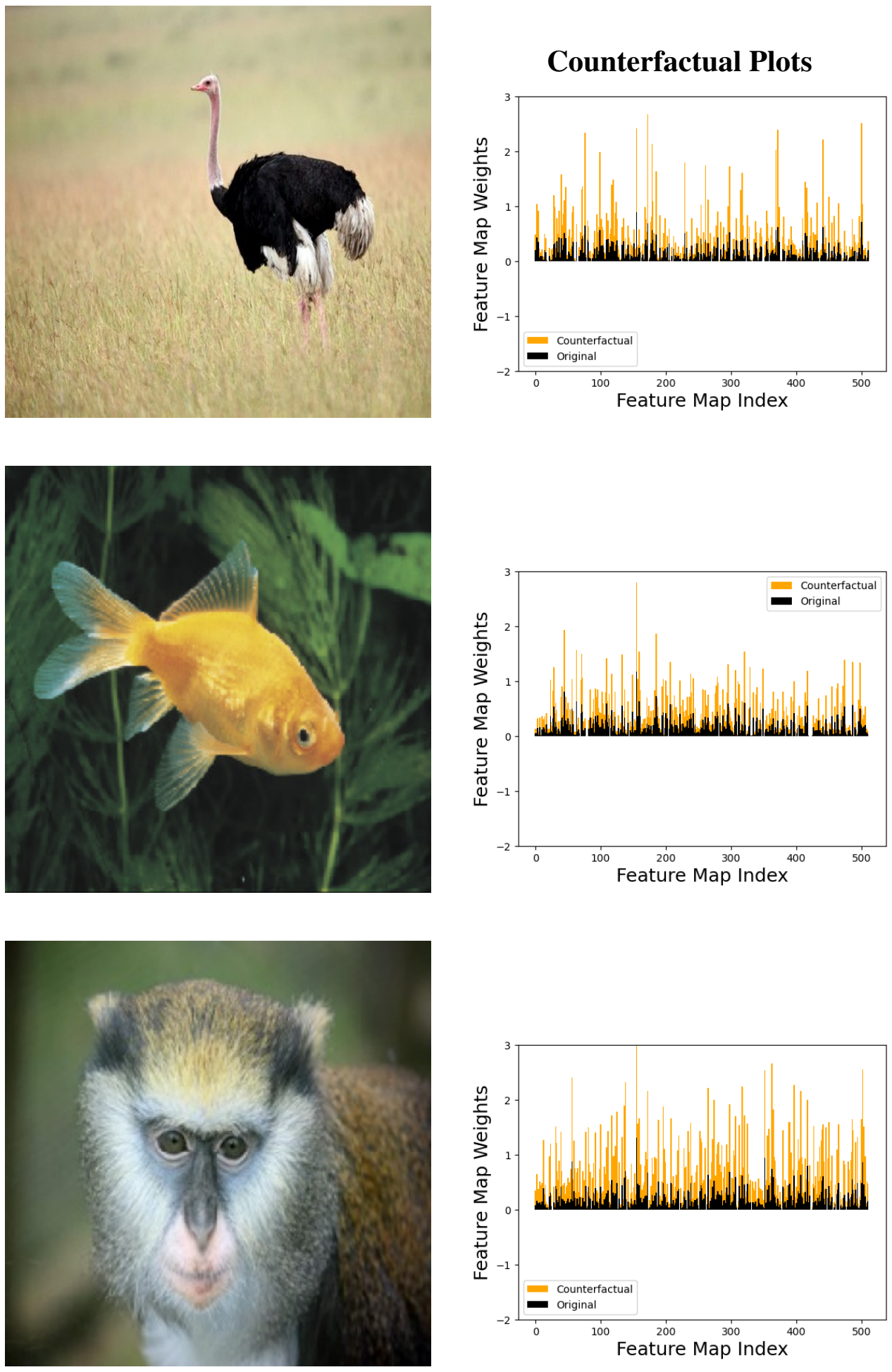

Figure 7. Graphical representation of counterfactual analysis results for verifying causal explanations. High variation between the actual and counterfactual mutual information scores show good causality.

### 4.6. Sanity Check

Sanity check has emerged as a standard means of assessing the sensitivity of CAM approaches towards model parameter randomization. Initially introduced in [31] , it evaluates the sanity of the approach, giving deeper insights apart from visual assessment. Cascading model parameter randomization test was deployed in the present work to evaluate the sanity of MI CAM. The results obtained by subsequent randomization of parameters from shallow layers to the deeper layers are summarized in Fig 8.

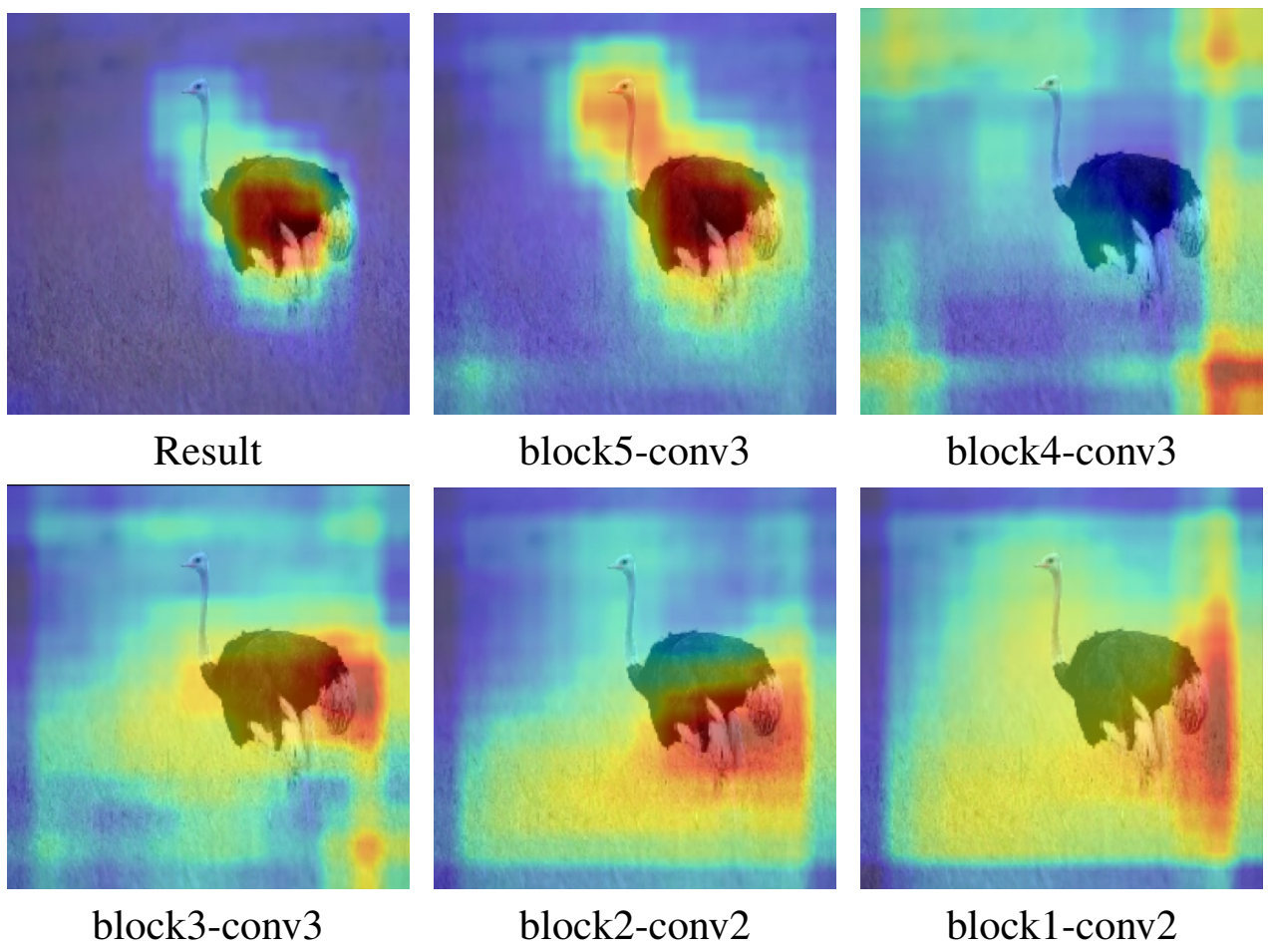

Figure 8. Sanity check for MI CAM of last convolutional layer of VGG-16 model

It can be inferred how parameter randomization disperses the saliency map. This shows the sensitivity of MI CAM to model parameters.

## 5. Ablation Study

We demonstrate the robustness of MI CAM against the issue of noisy feature map weights encountered in previous CAM methods by conducting weighting component visualizations and counterfactual comparison analysis. Detailed results of these experiments are provided in the supplementary material.

## 6. Conclusion and Future Direction

In this paper, we proposed MI-CAM a novel class-agnostic visual explanation approach for robust and precise explanations of CNNs. MI-CAM is deprived of class-based biases and shattered weights problems. We verify the causality in the generated explanations through counterfactual analysis and perform sanity check as well. We further theorize the need to examine only the final layer of a CNN as it encapsulates all information of the previous layers as per conditional independence. This can have many applications in localization and weakly supervised segmentation tasks, helping researchers venture further into energy based modeling and analysis of deep learning models and encoders for causal explanations. Further study can be done with the focus on reducing the computational cost and inference time of calculating mutual information in high dimensional data, methods like Hilbert-Schmidt independence criterion (HSIC) could be tested as potential cost efficient replacements for the traditional calculation of mutual information.

# MI CAM: Mutual Information Weighted Activation Mapping for Causal Visual Explanations of Convolutional Neural Networks

## Supplementary Material

## A. Ablation Study

Here we report the results of weighting component visualization and counterfactual comparative analysis in an attempt to showcase the robustness of MI CAM in generating feature map weights immune to noise and shattering.

### A.1. Weighting Component Visualization

Earlier CAM based visualization approaches such as gradient-based CAM approaches suffered the problem of noisy and unstable feature map weights due to the shattered gradient problem.

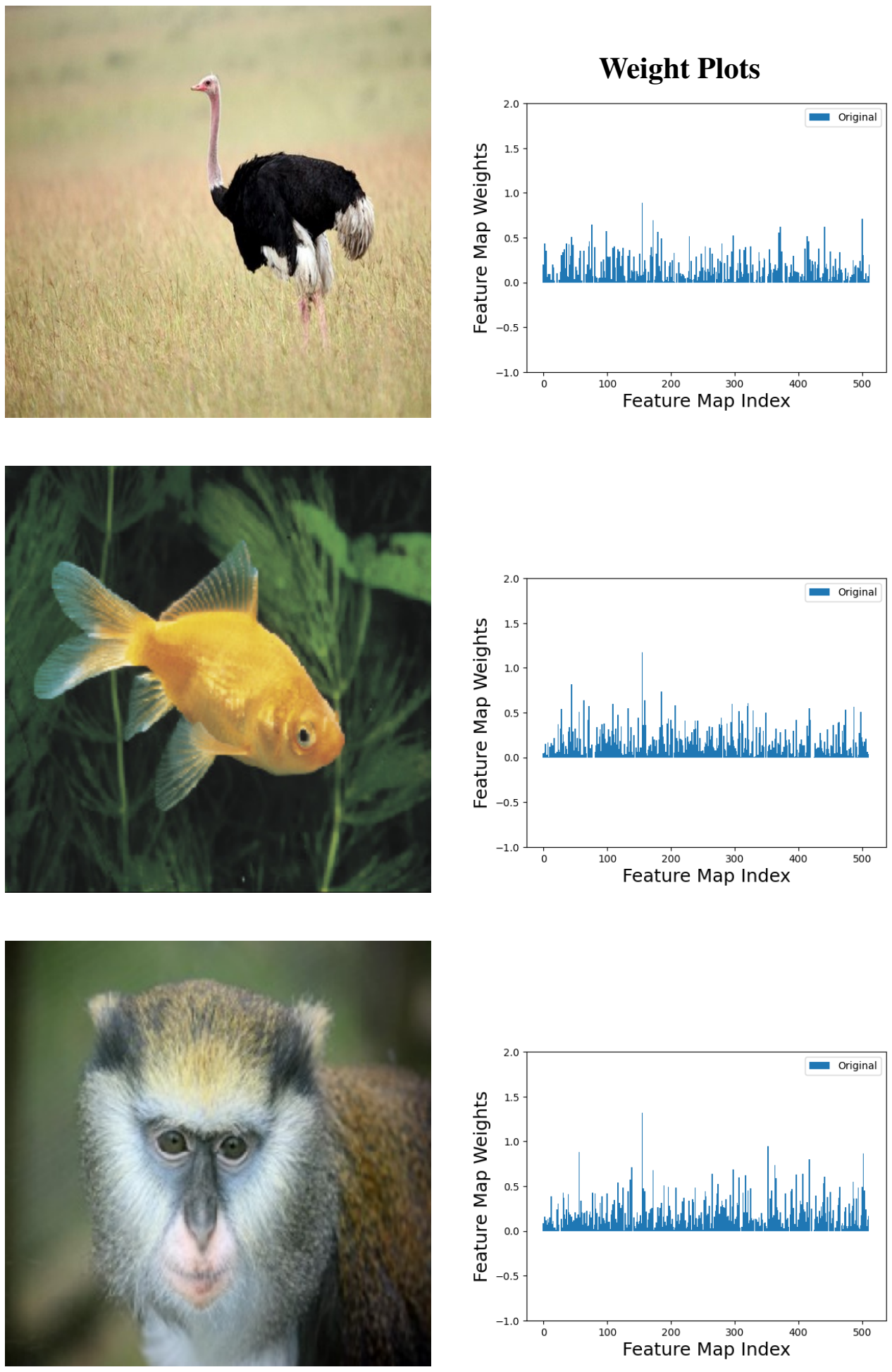

Figure 9. Weight component visualization

This makes the visualizations noisy and susceptible to errors. In order to validate the robustness of mutual information as a non-noisy and stable weighting criteria, we visualize the weighing components of MI CAM along the channels. The weighting components are showing stability and robustness towards noise in Fig 9. This validates for the use of mutual information between the feature maps and the input image as weighting parameter for each feature map in MI CAM.

### A.2. Comparative Causality Analysis

We reassert the importance of causality in the explanations and the corresponding counterfactual analysis discussed in our work by providing comparative studies of causalities in the explanations generated by earlier CAM approaches in this section. We provide the counterfactual plots of gradient-based and decomposition-based approaches.

It can be inferred from Fig 10 that both the earlier methods of gradient-based and decomposition-based CAM approaches are showing almost no sensitivity to counterfactual perturbations and hence no causality in their explanations. This further extends the practicality of MI CAM in producing causal explanations and the need for analysis of causality in the generated CAM in our study.

## B. Inference Speed Evaluation

As discussed in the paper earlier in Sec 1.1, higher amount of computational time and low inference speed is a major issue faced by decomposition and target class score based CAM approaches in comparison to other gradient-based approaches. It is also an issue attributed to the calculation of mutual information in high dimensional data like images and videos. However, though increase in performance comes with increased computational cost, MI CAM performs comparably to Score CAM, a state of the art decomposition-based CAM approach, with much less computational time and higher inference speed.

Table 4. Inference time comparison (in seconds)

| Model | Method | CPU | T4 GPU | TPU v2-8 |
|---|---|---|---|---|
| VGG-16 | Score CAM | 386.85 s | 13.58 s | 43.90 s |
| | MI CAM | **6.67 s** | **6.07 s** | **4.42 s** |
| MobileNet | Score CAM | 170.70 s | 75.44 s | 82.67 s |
| | MI CAM | **22.46 s** | **14.93 s** | **9.77 s** |

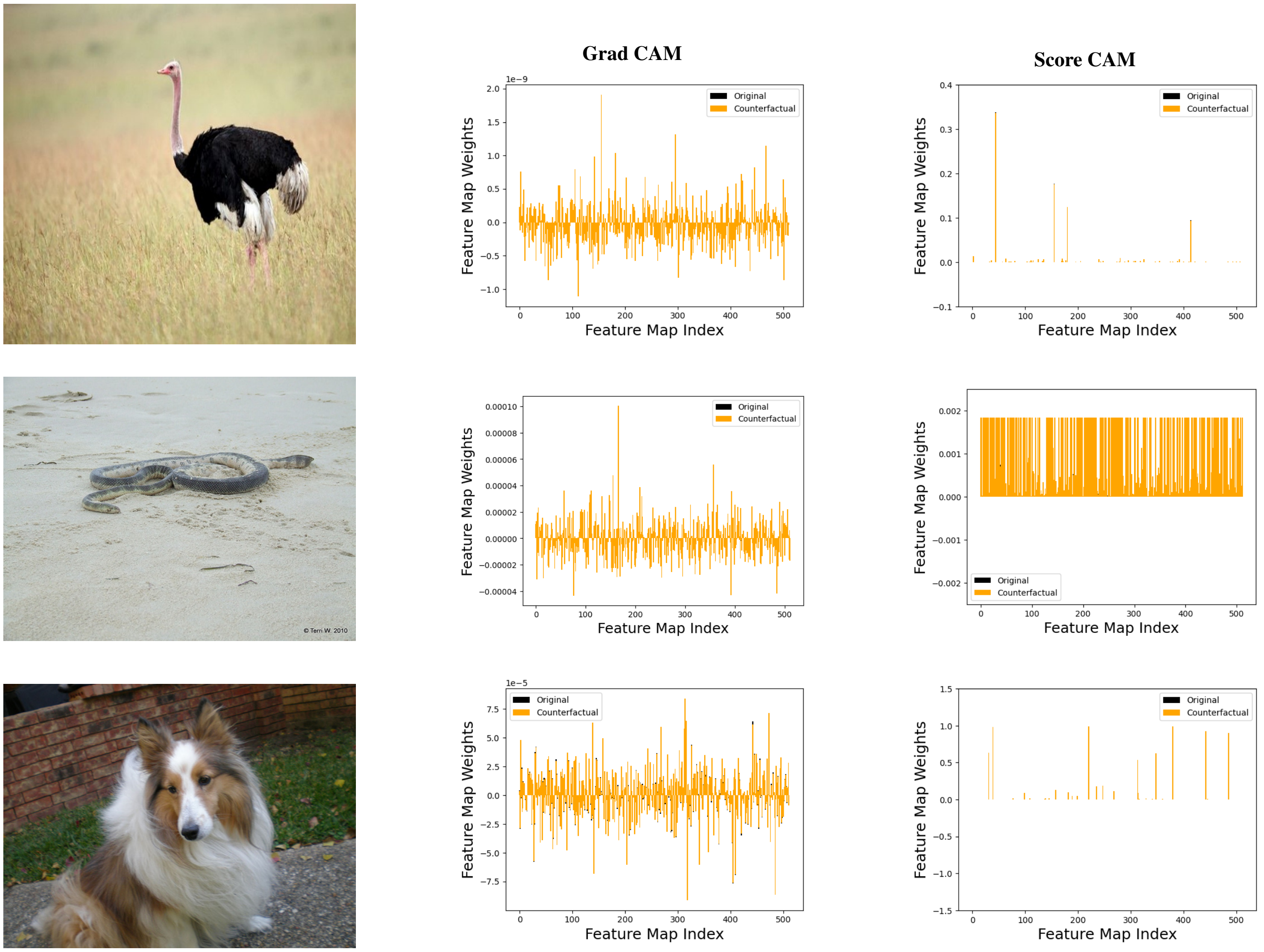

Figure 10. Comparative counterfactual plots of three instances from the ILSVRC2012 dataset for Grad CAM and Score CAM across channels. Plots show indistinguishably small sensitivity towards counterfactual changes.

Table 4 comprises of the inference time analysis results taken on VGG-16 and MobileNet for three devices, CPU, T4 GPU and TPU v2-8. Each result is the average taken over 10 runs on the sample. It can be inferred that MI CAM performs with 57× and 2× less inference time than Score CAM for CPU and T4 GPU respectively on VGG-16 and with 7.5× and 5× less inference time for CPU and T4 GPU respectively on MobileNet.

## C. Model Variation Analysis

To validate the performance of MI CAM, we analyze the quantitative evaluation of MI CAM on pre-trained MobileNet model through average drop (AD) and average increase (AI) metrics. Table 5 showcases the AD and AI results of MI CAM on MobileNet. The analysis is done on the ILSVRC2012 validation dataset.

Table 5. Evaluation results for AD and AI for MobileNet. Lower AD and Higher AI signifies a better performance of the explainability method.

| Method | Avg. Drop (AD) | Avg. Increase (AI) |
|---|---|---|
| Grad CAM | 88.637 | 45.733 |
| Grad CAM++ | 91.860 | 43.207 |
| Score CAM | 80.109 | 57.086 |
| XGrad CAM | 94.306 | 21.869 |
| Eigen CAM | 93.586 | 30.382 |
| MI CAM | **78.216** | **59.760** |

As shown, MI CAM achieves an AD of 78.21% and an AI of 59.76% on MobileNet, outperforming all other CAM approaches. This validates the generalizability of MI CAM across models of varying architecture.